\documentclass[runningheads]{llncs}
\usepackage[T1]{fontenc}
\usepackage{graphicx}
\usepackage{subcaption}
\usepackage{booktabs}
\usepackage[misc]{ifsym}
\newcommand{\corr}{(\Letter)}
\usepackage{hyperref}
\usepackage{amsfonts}
\usepackage{amsmath}
\usepackage{xcolor}
\usepackage{booktabs}   
\usepackage{multirow}   
\usepackage{rotating}   
\usepackage{array}      

\tocauthor{Furong Peng, Jinzhen Gao, Xuan Lu, Kang Liu, Yifan Huo, Sheng Wang}
\toctitle{Towards Deeper GCNs: Alleviating Over-smoothing via Iterative Training and Fine-tuning}

\usepackage{mwe}
\begin{document}

\title{Towards Deeper GCNs: Alleviating Over-smoothing via Iterative Training and Fine-tuning}
\titlerunning{Alleviating Over-smoothing via Iterative Training and Fine-tuning}


\author{Furong Peng\inst{1} \and
Jinzhen Gao\inst{2} \and
Xuan Lu\inst{3}\corr \and
Kang Liu\inst{4} \and
Yifan Huo\inst{5} \and
Sheng Wang\inst{6}}

\authorrunning{F. Peng et al.}

\institute{
Institute of Big Data Science and Industry, Shanxi University/Key Laboratory of Evolutionary Science Intelligence of Shanxi Province, Taiyuan, China
\email{pengfr@sxu.edu.cn},
\and
Shanxi University, Taiyuan, China
\email{202322404011@email.sxu.edu.cn},
\and
Shanxi University, Taiyuan, China
\email{xuanlu@sxu.edu.cn},
\and
Shanxi University, Taiyuan, China
\email{202222407023@email.sxu.edu.cn},
\and
Shanxi University, Taiyuan, China
\email{202422409007@email.sxu.edu.cn},
\and
Zhengzhou University of Aeronautics, Zhengzhou, China
\email{wangsheng1910@zua.edu.cn}
}


\maketitle              

\begin{abstract}
Graph Convolutional Networks (GCNs) suffer from severe performance degradation in deep architectures due to over-smoothing. While existing studies primarily attribute the over-smoothing to repeated applications of graph Laplacian operators, our empirical analysis reveals a critical yet overlooked factor: trainable linear transformations in GCNs significantly exacerbate feature collapse, even at moderate depths (e.g., 8 layers). In contrast, Simplified Graph Convolution (SGC), which removes these transformations, maintains stable feature diversity up to 32 layers, highlighting linear transformations' dual role in facilitating expressive power and inducing over-smoothing. However, completely removing linear transformations weakens the model's expressive capacity.

To address this trade-off, we propose \textbf{Layer-wise Gradual Training (LGT)}, a novel training strategy that progressively builds deep GCNs while preserving their expressiveness. LGT integrates three complementary components: (1) \textit{layer-wise training} to stabilize optimization from shallow to deep layers, (2) \textit{low-rank adaptation} to fine-tune shallow layers and accelerate training, and (3) \textit{identity initialization} to ensure smooth integration of new layers and accelerate convergence. Extensive experiments on benchmark datasets demonstrate that LGT achieves state-of-the-art performance on vanilla GCN, significantly improving accuracy even in 32-layer settings. Moreover, as a training method, LGT can be seamlessly combined with existing methods such as PairNorm and ContraNorm, further enhancing their performance in deeper networks. LGT offers a general, architecture-agnostic training framework for scalable deep GCNs. The code is available at [\url{https://github.com/jfklasdfj/LGT_GCN}].

\keywords{GCNs  \and Over-smoothing \and Fine-tune  \and LoRA}
\end{abstract}

\section{Introduction}

Graph Neural Networks (GNNs) \cite{defferrard2016convolutional,feng2020graph,gilmer2017neural} have become a powerful paradigm for learning from graph-structured data, achieving notable success across applications such as social network analysis \cite{fan2019graph,zhou2020graph}, molecular property prediction \cite{lin2020kgnn}, and traffic forecasting \cite{li2024survey}. Among them, Graph Convolutional Networks (GCNs) \cite{kipf2017semisupervised,chen2020simple} are widely adopted for their ability to aggregate neighborhood information through iterative message passing. 

\begin{figure}[htbp]
	\centering
	\begin{subfigure}[b]{0.48\textwidth}
		\includegraphics[width=\textwidth]{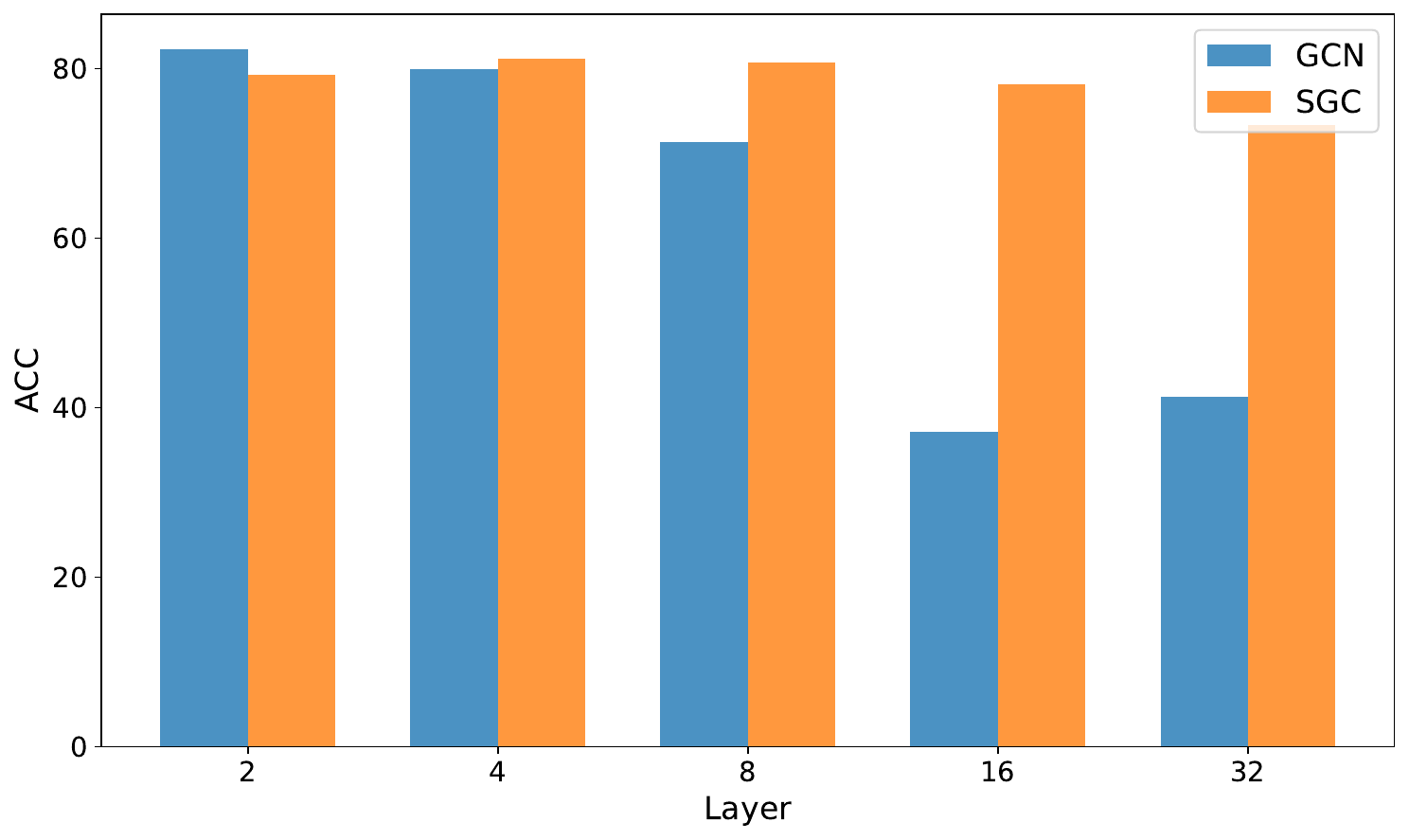} %
		\caption{Accuracy comparison of SGC and GCN at different depths. }
		\label{fig:left}
	\end{subfigure}
	\hfill
	\begin{subfigure}[b]{0.48\textwidth}
		\begin{subfigure}[b]{0.48\textwidth}
			\includegraphics[width=\textwidth]{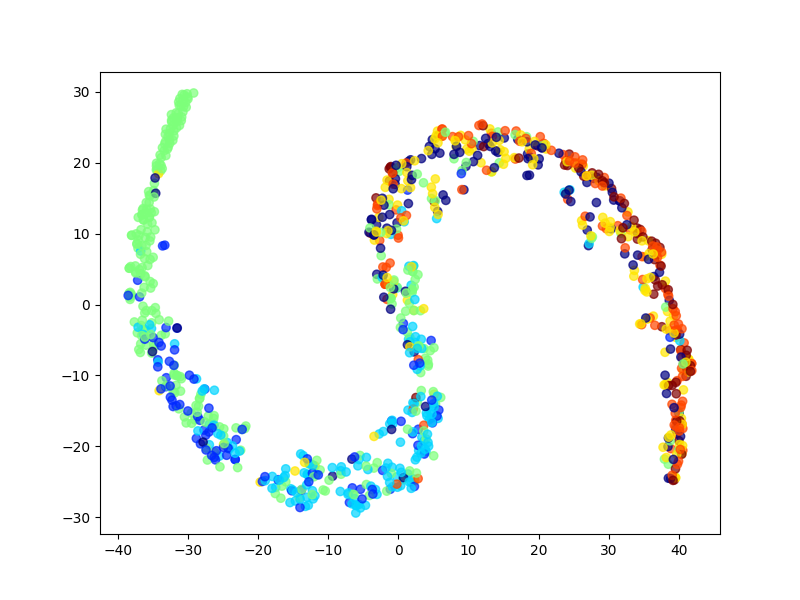} 
			\caption{ GCN (8 layers)}
			\label{fig:top-left}
		\end{subfigure}
		\hfill
		\begin{subfigure}[b]{0.48\textwidth}
			\includegraphics[width=\textwidth]{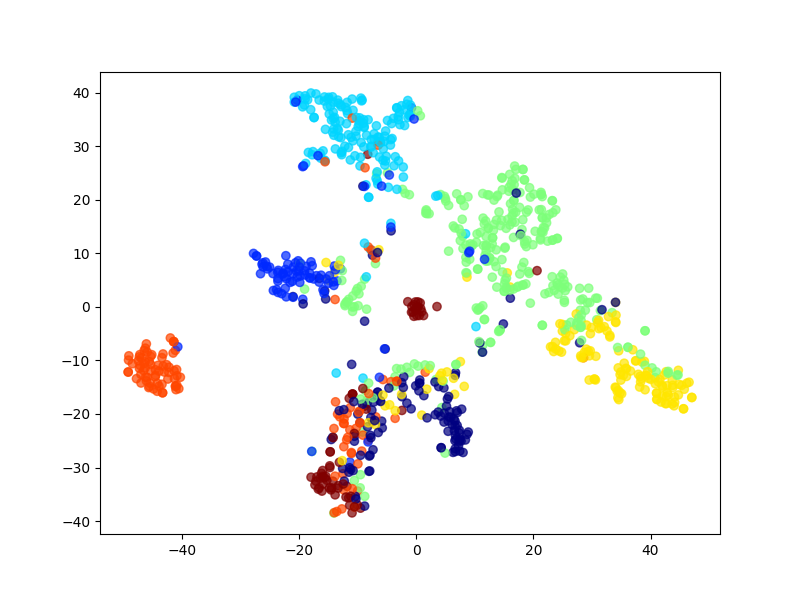} 
			\caption{SGC (8 layers)}
			\label{Top Right Subfigure}
		\end{subfigure}
		
		\vspace{\floatsep} 
		
		\begin{subfigure}[b]{0.48\textwidth}
			\includegraphics[width=\textwidth]{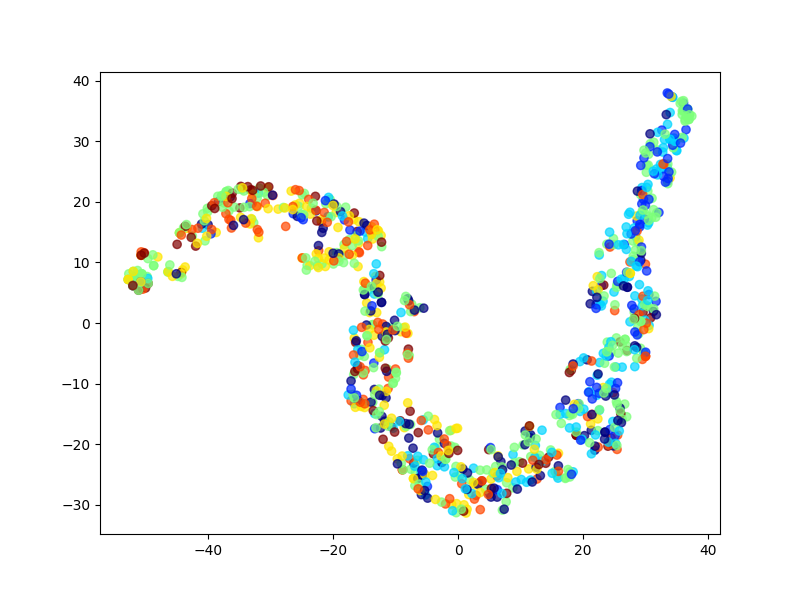} 
			\caption{GCN (32 layers)}
			\label{fig:bottom-left}
		\end{subfigure}
		\hfill
		\begin{subfigure}[b]{0.48\textwidth}
			\includegraphics[width=\textwidth]{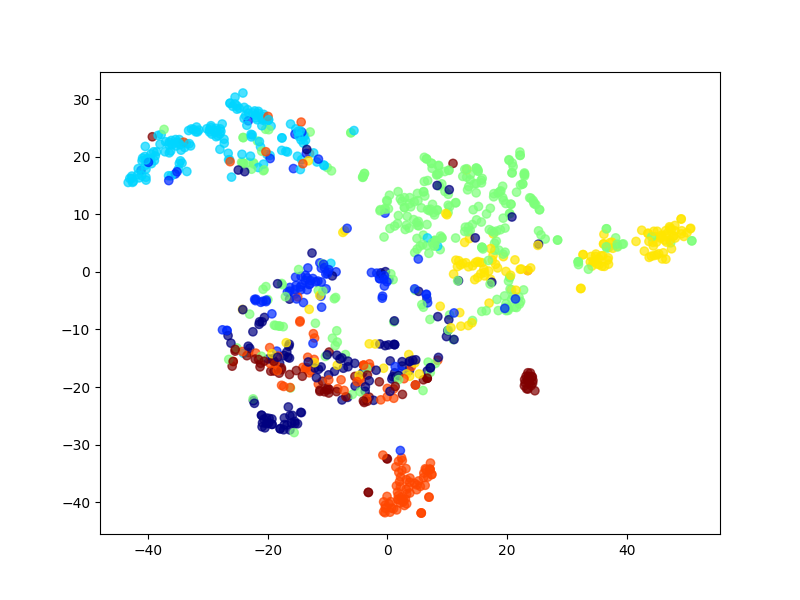} 
			\caption{SGC (32 layers)}
			\label{fig:bottom-right}
		\end{subfigure}
	\end{subfigure}
	\caption{\textbf{Comparison of over-smoothing in GCN and SGC on Cora.} (a) Accuracy trend with increasing depth, where GCN suffers severe degradation beyond 8 layers, while SGC maintains stable performance. (b-e) Node embedding visualization (via t-SNE \cite{van2008tsne}) at different depths: GCN's features collapse as depth increases, whereas SGC preserves clear class separability even at 32 layers.}
	\label{fig:over-smoothing-analysis}
\end{figure}

Despite their effectiveness, GCNs face severe performance degradation when scaled to deeper architectures due to the over-smoothing problem, where node representations become indistinguishable across layers \cite{Oono2019GraphNN,chen2020measuring}. Although previous studies primarily attribute over-smoothing to repeated applications of the graph Laplacian\cite{li2018deeper}, our empirical analysis (Figure~\ref{fig:over-smoothing-analysis}) reveals a critical yet under-explored cause: the linear transformations in GCN layers substantially accelerate feature collapse, even at moderate depths such as 8 layers. Interestingly, as shown in Figure~\ref{fig:over-smoothing-analysis}, Simplified Graph Convolution (SGC) \cite{wu2019simplifying}, which removes both linear transformations and nonlinearities, maintains stable performance and separable node representations even at 32 layers. This contrast highlights a fundamental trade-off: while linear transformations are essential for expressive feature learning, they also intensify over-smoothing in deep GCNs. Existing methods, however, largely overlook this dual role, leaving the challenge of balancing expressiveness and smoothness unresolved.

To address the limitations of deep GCNs, we introduce Layer-wise Gradual Training (LGT), a novel training paradigm designed to mitigate over-smoothing while preserving model expressiveness. Instead of training all layers simultaneously, LGT progressively expands the network depth, integrating three key strategies: incremental layer-wise training, low-rank adaptation (LoRA), and identity initialization.

\textbf{Incremental Layer-wise Training}. Traditional GCNs perform well in shallow settings but deteriorate rapidly as depth increases. To leverage this property, LGT adopts a staged training approach, gradually increasing depth while ensuring node embeddings remain discriminative at each step. Unlike prior layer-wise training strategies that require retraining the entire model upon adding new layers, LGT maintains previously learned representations, reducing computational redundancy and improving convergence stability.

\textbf{Low-Rank Adaptation for Efficient Fine-tuning}. While freezing earlier layers prevents over-smoothing, it also limits adaptation to new layers. To address this, LGT integrates Low-Rank Adaptation (LoRA) \cite{hu2022lora}, enabling lightweight fine-tuning of pre-trained layers via a low-rank decomposition of weight updates. This approach significantly accelerates convergence, reducing the need for full retraining while maintaining representational flexibility.

\textbf{Identity Initialization for Stable Expansion}. Random initialization of newly added layers can disrupt learned representations, causing performance fluctuations. Inspired by SGC \cite{wu2019simplifying}, which maintains performance using a fixed identity matrix as the linear transformation, LGT initializes new layers using identity matrices. This ensures smoother integration, reducing optimization instability and enhancing training efficiency.

By combining these strategies, LGT enables deep GCNs to achieve superior depth scalability, faster convergence, and improved classification accuracy while maintaining a lightweight and architecture-agnostic design. This framework not only provides a novel solution to over-smoothing but also establishes a robust foundation for training deep graph networks efficiently. Extensive experiments on semi-supervised node classification benchmarks demonstrate that LGT significantly alleviates over-smoothing and achieves state-of-the-art (SOTA) performance in deep vanilla GCNs (e.g., 32 layers), outperforming existing anti-over-smoothing methods. Furthermore, LGT is model-agnostic and can be seamlessly integrated with normalization-based techniques (e.g., PairNorm \cite{zhao2020pairnorm} and ContraNorm \cite{guo2023contranorm}) to further improve their performance.

In summary, our contributions are as follows:
\begin{itemize}
	\item We identify trainable linear transformations as a critical cause of over-smoothing in moderate depth and propose a \textit{training-based solution} fundamentally different from prior architectural modifications or regularization approaches.
	\item We introduce Layer-wise Gradual Training (LGT), a general and efficient strategy that combines incremental training, low-rank adaptation, and identity initialization, substantially improving the performance of deep vanilla GCNs and achieving SOTA results. LGT is also compatible with existing anti-over-smoothing techniques.
\end{itemize}

The remainder of this paper is organized as follows: Section~\ref{sec:related_work} reviews related work, Section~\ref{sec:problem} formalizes the problem to evaluate over-smoothing, Section~\ref{sec:method} details the proposed LGT method, Section~\ref{sec:experiments} presents experimental results, and Section~\ref{sec:conclusion} concludes the paper.

\section{Related Work}\label{sec:related_work}

Existing approaches to alleviating over-smoothing in GCNs can be broadly categorized into two groups: (1) modifying message-passing mechanisms and (2) constraining node representations. Below, we review models in each category.

\subsection{Message-Passing-Based Approaches}

Message-passing-based methods aim to adjust the information aggregation process to mitigate over-smoothing \cite{peng2024tsc,wang2023snowflake}.  Graph Attention Networks (GAT) \cite{velivckovic2017graph} assign adaptive weights to neighbors, promoting selective aggregation. However, GAT models often face gradient instability and complex optimization in deep architectures. DropEdge \cite{Rong2020DropEdge} randomly removes edges during training, reducing redundant aggregation. OrderedGNN \cite{song2023ordered} structurally organizes neurons based on hop distances to constrain message passing. ResGCN \cite{li2019deepgcns} and JK-Net \cite{xu2018representation} introduce cross-layer connections to retain information from shallow layers, while PSNR \cite{zhou2024deep} combines residual links with adaptive normalization to capture multi-hop features. Although effective to some extent, these methods rely heavily on graph-dependent heuristics or sensitive hyperparameter tuning, limiting their generalization. 

\subsection{Representation-Based Approaches}

Another line of work focuses on directly constraining node representations to prevent feature homogenization. EGNN \cite{zhou2021dirichlet} introduces Dirichlet energy regularization to preserve informative gradients across layers. ContraNorm \cite{guo2023contranorm} and PairNorm \cite{zhao2020pairnorm} maintain feature variance through normalization, while BatchNorm-GCN \cite{chen2022learning} applies batch normalization to stabilize feature distributions. Although these methods effectively mitigate over-smoothing, they may also suppress useful feature interactions when applied excessively, limiting model expressiveness.

\subsection{Our Motivation and Distinction}

In contrast to existing methods that focus on modifying message passing or enforcing explicit feature constraints, our proposed LGT addresses over-smoothing from a training strategy perspective. By incrementally training deeper layers and applying low-rank adaptation to earlier layers, LGT implicitly regularizes feature propagation without altering GCN architecture or graph structure. Unlike prior methods, LGT maintains representation diversity and depth scalability through an optimization-centered approach, and is inherently compatible with normalization-based techniques such as PairNorm and ContraNorm for further performance gains.

\section{Problem Definition}\label{sec:problem}

Referring the methods of evaluating over-smoothing \cite{kipf2017semisupervised,peng2024tsc}, we consider the standard semi-supervised node classification task on a graph \( \mathcal{G} = (\mathcal{V}, \mathcal{E}) \), where \( \mathcal{V} = \{v_1, v_2, \dots, v_n\} \) is the set of nodes and \( \mathcal{E} \) is the set of edges. Each node \( v_i \in \mathcal{V} \) is associated with a feature vector \( x_i \in \mathbb{R}^f \), and all node features form the feature matrix \( \textbf{X} \in \mathbb{R}^{n \times f} \). A subset of nodes \( \mathcal{V}_L = \{v_1, \dots, v_m\} \) with \( m \ll n \) are labeled, while the remaining nodes \( \mathcal{V}_U = \mathcal{V} \setminus \mathcal{V}_L \) are unlabeled.

The graph structure is encoded via the adjacency matrix \( \textbf{A} \in \{0,1\}^{n \times n} \), where \( \textbf{A}_{ij} = 1 \) if an edge exists between nodes \( v_i \) and \( v_j \). We adopt the normalized graph Laplacian:
\begin{equation}
	\textbf{L }= \textbf{D}^{-\frac{1}{2}}(\textbf{A} + \textbf{I}) \textbf{D}^{-\frac{1}{2}},
\end{equation}
where \( \textbf{I} \) is the identity matrix and \( \textbf{D} \) is the diagonal degree matrix with \( \textbf{D}_{ii} = \sum_j \textbf{A}_{ij} \).

The objective of semi-supervised node classification is to train a GCN that minimizes the cross-entropy loss over labeled nodes:
\begin{equation}
	\mathcal{L} = - \sum_{v_i \in Y_L} \sum_{c=1}^{C} y_{i,c} \log \hat{y}_{i,c},
\end{equation}
where \( C \) is the number of classes, \( y_{i,c} \) is the one-hot label, and \( \hat{y}_{i,c} \) is the predicted class probability for node \( v_i \).

A key challenge arises when scaling GCNs to deep architectures due to the over-smoothing effect, where node representations \( H^{(K)} \) progressively converge to a subspace with minimal discriminative information:
\begin{equation}
	\lim_{K \rightarrow \infty} H^{(K)} \approx \mathbf{C},
\end{equation}
where \( \mathbf{C} \) is a constant matrix independent of node identity. This phenomenon severely degrades classification accuracy on unlabeled nodes. Therefore, an effective GCN should maintain high classification accuracy even as the network depth \( K \) increases. Our goal is to design a training strategy that mitigates over-smoothing while enabling deeper GCNs to preserve discriminative node representations.

\section{Layer-wise Gradual Training}
\label{sec:method}

While deep GCNs suffer from over-smoothing, our empirical analysis (Figure~\ref{fig:over-smoothing-analysis}) reveals that trainable linear transformations accelerate feature collapse even at moderate depths. However, removing these transformations compromises model expressiveness, highlighting a key trade-off: GCNs require transformation layers for feature learning but must mitigate their role in over-smoothing to remain effective in deep architectures.

\begin{figure}
	\centering
	\includegraphics[width=0.7\textwidth]{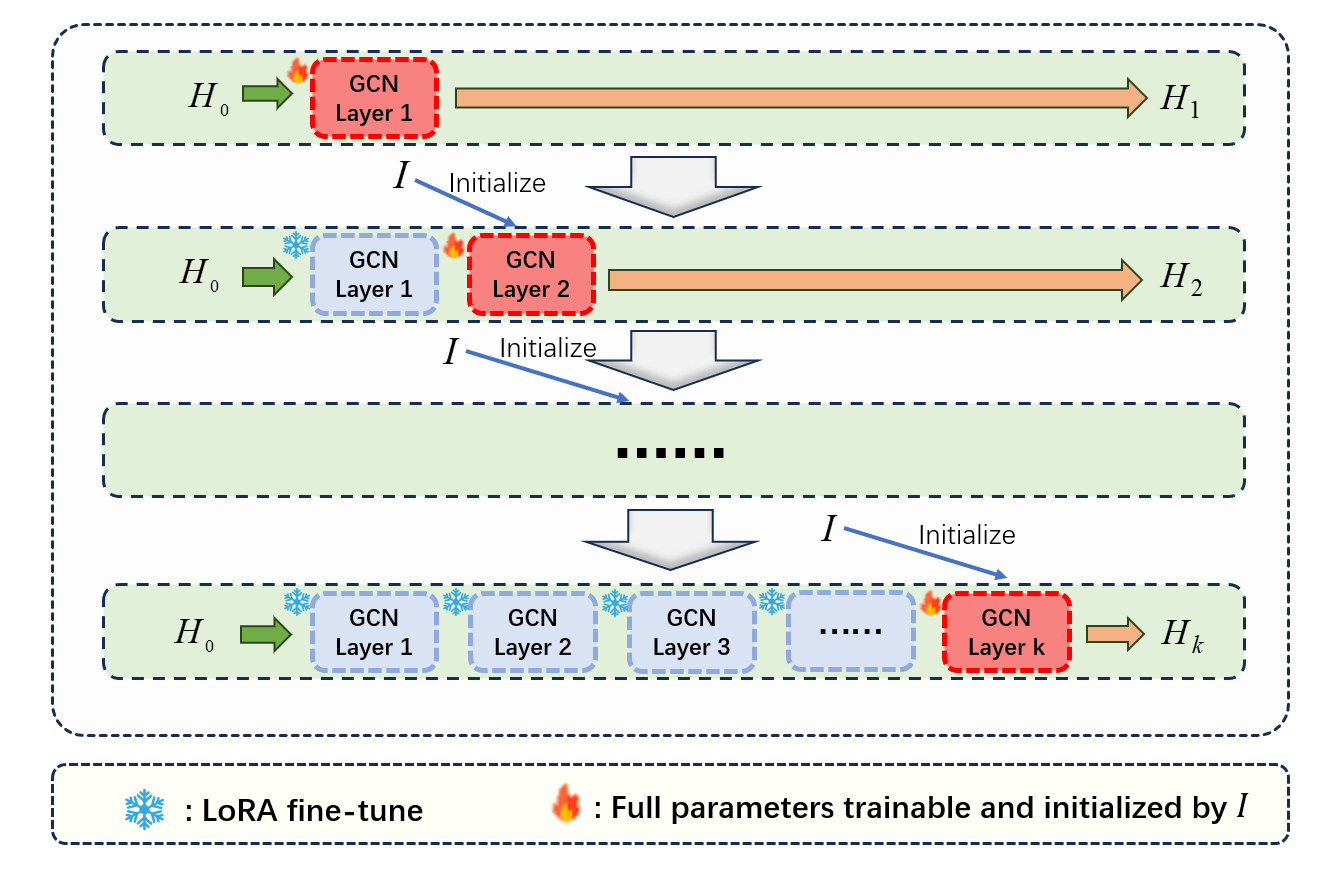}
	\caption{\textbf{Illustration of Layer-wise Gradual Training (LGT).} At each stage, only the newly added layer is fully trained, while shallow layers are fine-tuned via LoRA. Once stabilized, a new layer is added and initialized with an identity matrix to ensure smooth integration.}
	\label{fig:lgt}
\end{figure}

To address this, we introduce Layer-wise Gradual Training (LGT)—a progressive training strategy that incrementally deepens GCNs while preserving feature discriminability at each stage, as shown in Figure~\ref{fig:lgt}. Instead of training all layers simultaneously, LGT stabilizes optimization and prevents feature degradation through structured training. LGT consists of three key components: (1) \textit{Incremental Layer-wise Training}, which prevents abrupt optimization difficulties by progressively adding layers, ensuring stable feature learning at each depth; (2) \textit{Low-Rank Adaptation (LoRA)} \cite{hu2022lora}, which lightly fine-tunes frozen layers to retain expressiveness without excessive parameter updates. (3) \textit{Identity Matrix Initialization}, which ensures smooth integration of new layers, reducing parameter shifts and accelerating convergence. Empirical results (Section~\ref{sec:experiments}) confirm that LGT significantly improves classification accuracy in deep GCNs (e.g., 32 layers) while reducing computational overhead. The following sections detail each component.

\subsection{Incremental Layer-wise Training}

To address over-smoothing caused by rapid parameter growth, LGT decomposes GCN training into multiple stages. At each stage, only one new layer is added and trained, while all previously trained layers are fine-tuned (via LoRA). This staged optimization prevents feature collapse between layers and stabilizes feature propagation as depth increases.

\textbf{First Layer Training.} We start by training the first GCN layer to capture local neighborhood structures. Training proceeds until convergence based on validation performance, after which the layer is frozen.

\textbf{Incremental Layer Addition.} New GCN layers are added sequentially. At each step, only the new layer is fully
 trained, while shallow layers are slightly adapted using LoRA. Early stopping is applied to each stage to prevent overfitting. This procedure ensures that each layer progressively refines representations without causing over-smoothing or collapse.

\subsection{Low-Rank Adaptation (LoRA)}

While freezing earlier layers stabilizes training, it may restrict representation learning capacity. To balance stability and flexibility, we introduce LoRA \cite{hu2022lora} to adaptively fine-tune frozen layers using low-rank updates. Instead of fully updating the weight matrix \( \textbf{W}^{(k)} \in \mathbb{R}^{d_{\text{in}} \times d_{\text{out}}} \), we add a learnable low-rank component:
\begin{equation}
	\tilde{\textbf{W}}^{(k)} = \textbf{W}_0^{(k)} + \textbf{A}^{(k)} \textbf{B}^{(k)},
\end{equation}
where \( \textbf{W}_0^{(k)} \) is the frozen pretrained weight, and \( \textbf{A}^{(k)} \in \mathbb{R}^{d_{\text{in}} \times r}, \textbf{B}^{(k)} \in \mathbb{R}^{r \times d_{\text{out}}} \) are low-rank matrices with \( r \ll \min(d_{\text{in}}, d_{\text{out}}) \). The forward propagation becomes:
\begin{equation}
	\textbf{H}^{(k+1)} = \sigma( \textbf{L} \textbf{H}^{(k)} (\textbf{W}_0^{(k)} + \textbf{A}^{(k)} \textbf{B}^{(k)})),
\end{equation}
and only \( \textbf{A}^{(k)} \), \( \textbf{B}^{(k)} \) are updated during training. This significantly reduces memory and computational cost while preserving adaptability in deep models.

\subsection{Identity Matrix Initialization}

\begin{figure}[t]
	\centering
	\begin{subfigure}[b]{0.48\textwidth}
		\includegraphics[width=\textwidth, height=5.2cm]{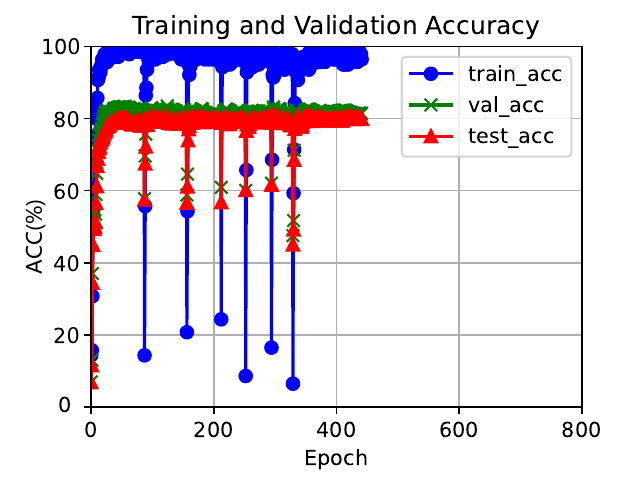}
		\caption{Random Initialization.}
		\label{fig:lt-wo-identity}
	\end{subfigure}
	\centering
	\begin{subfigure}[b]{0.48\textwidth}
		\includegraphics[width=\textwidth, height=5.2cm]{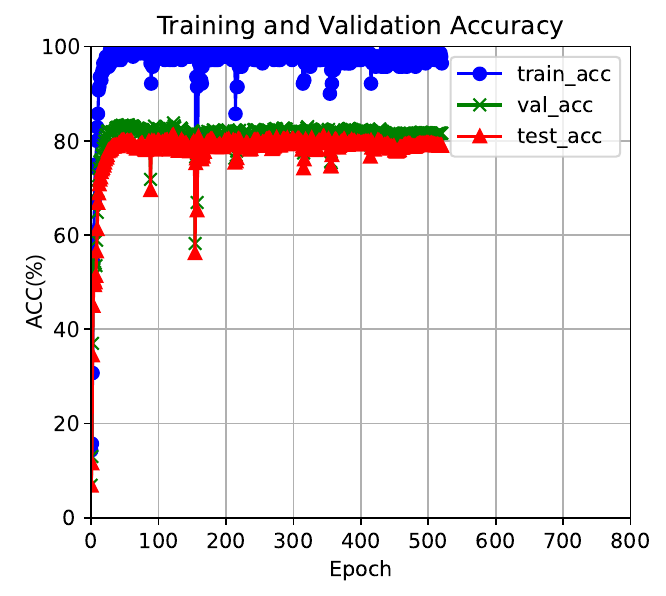}
		\caption{Identity Initialization.}
		\label{fig:lt-identity}
	\end{subfigure}
	\caption{\textbf{Effect of initialization methods.} Identity initialization stabilizes the training of newly added layers compared to random initialization.}
	\label{fig:identity}
\end{figure}
Proper initialization of new layers is crucial for stable incremental training. Random initialization may lead to unstable gradients and disrupt previously learned representations, as shown in Figure~\ref{fig:identity}. We initialize each newly added layer's weights as an identity matrix, ensuring initial outputs align with the prior layer's features:
\begin{equation}
	\textbf{H}^{(k+1)} = \sigma( \textbf{L} \textbf{H}^{(k)} \textbf{W}^{(k)} ) \approx \sigma( \textbf{L} \textbf{H}^{(k)} ),
\end{equation}
thus providing a smooth transition for training deeper layers. As demonstrated in Figure~\ref{fig:identity}, identity initialization significantly improves stability and accelerates convergence compared to random initialization.

\subsection{Summary}

In summary, LGT integrates incremental layer-wise training, low-rank adaptation, and identity initialization into a unified framework that enables deep GCNs to avoid over-smoothing while retaining strong representational power. Unlike prior methods that modify GCN architectures or impose explicit constraints on node representations, LGT addresses over-smoothing through an optimization-centric training paradigm, offering a scalable and compatible solution for training deep GCNs. The effectiveness of LGT is validated through extensive experiments in the next section.

\section{Experiments}\label{sec:experiments}

We conduct comprehensive experiments to evaluate the effectiveness of our proposed LGT in alleviating over-smoothing for deep GCNs on \textit{semi-supervised node classification} tasks \cite{kipf2017semisupervised}. All models are evaluated using classification accuracy (ACC) as the primary metric. Our evaluation covers 4 benchmark datasets and 6 state-of-the-art baselines. 

\subsection{Experimental Setup}
\paragraph{Datasets.}
To evaluate the generalizability of LGT across diverse graph domains, we conduct experiments on 4 widely used citation and co-authorship datasets: Cora \cite{mccallum2000automating,sen2008collective}, Citeseer \cite{giles1998citeseer,sen2008collective}, Pubmed \cite{namata2012query,sen2008collective}, and AmazonPhoto \cite{shchur2018pitfalls}. The statistics of these datasets are summarized in Table~\ref{tb:dataset}. 
	\begin{table}[h]
	\centering
	\caption{The summary  of dataset}
	\begin{tabular}{llrrrr}
		\hline
		Datasets    & Cls. & Nodes & Edges  & Feat. & Train/Valid/Test \\ \hline
		Cora        & 7    & 2708  & 5429   & 1433  & 140/1000/1000    \\
		Citeseer    & 6    & 3327  & 4732   & 2703  & 120/1000/1000    \\
		Pubmed      & 3    & 19717 & 44338  & 500   & 60/1000/1000     \\
		AmazonPhoto & 8    & 7650  & 119043 & 745   & 160/1000/1000    \\\hline
	\end{tabular}
	\label{tb:dataset}
\end{table}
For all datasets, we follow the widely adopted semi-supervised learning setting \cite{kipf2017semisupervised}, randomly selecting 20 labeled nodes per class for training, and reserving 1,000 nodes each for validation and test sets. The remaining nodes are treated as unlabeled. To ensure robustness, we repeat each experiment with 5 random splits and report the mean performance.

\paragraph{Baselines.}
To thoroughly validate the effectiveness of LGT, we compare several representative and state-of-the-art GCN-based models that address over-smoothing. The baselines include:

\begin{itemize}
	\item \textbf{GCN (ICLR'17)} \cite{kipf2017semisupervised}: A foundational graph convolutional network model that performs neighborhood aggregation with learned linear transformations.
	\item \textbf{SGC (ICML'19)} \cite{wu2019simplifying}: A simplified GCN variant that removes nonlinear activations and collapses multiple layers into a single linear transformation to study the role of propagation.
	\item \textbf{PairNorm (ICLR'20)} \cite{zhao2020pairnorm}: A normalization technique that preserves feature variance across layers to prevent feature collapse.
	\item \textbf{ContraNorm (ICLR'23)} \cite{guo2023contranorm}: A method introducing implicit feature de-correlation to maintain node diversity and alleviate over-smoothing.
	\item \textbf{IresGCN (ICML'24)} \cite{park2024mitigating}: An inverse residual GCN framework that adjusts message passing directions to preserve personalized node information.
	\item \textbf{PSNR (NeurIPS'24)} \cite{zhou2024deep}: A posterior sampling and adaptive residual method that adaptively retains hierarchical information during propagation.
\end{itemize}

These baselines cover a wide range of anti-over-smoothing techniques, including normalization, residual connections, and structural adjustments. Thus, they provide a strong basis for assessing the performance and compatibility of LGT.

\subsection{Effectiveness on Alleviating Over-smoothing}
\definecolor{improvement-color}{RGB}{0, 102, 204} 
\definecolor{best-color}{RGB}{255, 0, 0}          
\definecolor{second-best-color}{RGB}{0, 200, 0}  
\newcommand{\improvement}{\textcolor{improvement-color}{$^{\uparrow}$}} 
\newcommand{\best}[1]{\textcolor{best-color}{#1}}
\newcommand{\secondbest}[1]{\textcolor{second-best-color}{#1}}

To evaluate LGT’s effectiveness in mitigating over-smoothing, we measure classification accuracy across different depths (4, 8, 16, and 32 layers) on four benchmark datasets: Cora, Citeseer, Pubmed, and AmazonPhoto. LGT is applied to GCN, ContraNorm, and PairNorm to assess its generalizability. The complete results are shown in Table~\ref{master-table}, where the performances of \best{best} and \secondbest{second-best} are highlighted, and the improvements brought about by LGT are marked with \improvement.

\begin{table}[!t]
	\centering
	\caption{Classification accuracy (ACC) across different datasets (in percent)}
	\begin{tabular}{c@{\hspace{1em}}lllll}
		\toprule
		\multirow{2}{*}{\centering Dataset} & \multicolumn{1}{c}{\multirow{2}{*}{ Model}}  & \multicolumn{4}{c}{Layers} \\
		\cmidrule(lr){3-6}
		 & & 4 & 8 & 16 & 32 \\
		\midrule
		\multirow{10}{*}{\rotatebox[origin=c]{90}{Cora}}
		& SGC & $79.88_{\pm 1.00}$ & $79.08 _{\pm 1.42}$ & $75.90 _{\pm 1.91}$ & $69.62 _{\pm 2.57}$ \\
		& IresGCN & $79.28 _{\pm 1.85}$ & $78.78 _{\pm 0.63}$ & $79.04 _{\pm 0.54}$ & $78.94 _{\pm 0.92}$ \\
		& PSNR & $79.90 _{\pm 1.00}$ & $79.00 _{\pm 1.39}$ & $79.54 _{\pm 0.73}$ & $79.48 _{\pm 0.83}$ \\\cline{2-6}
		& GCN & $78.58 _{\pm 1.29}$ & $71.30 _{\pm 1.54}$ & $37.20 _{\pm 2.22}$ & $40.46 _{\pm 0.78}$ \\
		& GCN+LGT & \best{$80.94_{\pm 1.03}$}\improvement & \best{$80.76_{\pm 1.28}$}\improvement & \secondbest{$80.58_{\pm 0.49}$}\improvement & \best{$81.06 _{\pm 0.97}$}\improvement \\\cline{2-6}
		& ContraNorm & $79.74 _{\pm 0.77}$ & $78.96 _{\pm 1.06}$ & $79.42 _{\pm 1.88}$ & $80.10 _{\pm 1.01}$ \\
		& ContraNorm+LGT & \secondbest{$80.36 _{\pm 1.20}$}\improvement & \secondbest{$80.34 _{\pm 0.94}$}\improvement & \best{$80.66 _{\pm 1.63}$}\improvement & \secondbest{$80.62 _{\pm 0.84}$}\improvement \\\cline{2-6}
		& PairNorm & $78.40 _{\pm 1.88}$ & $78.10 _{\pm 1.05}$ & $77.76 _{\pm 1.70}$ & $77.46 _{\pm 1.22}$ \\
		& PairNorm+LGT & $79.50_{\pm 0.56}$\improvement & $77.50_{\pm 1.49}$ & $78.88 _{\pm 0.24}$\improvement & $79.68 _{\pm 1.33}$\improvement \\
		
		\toprule
		
		\multirow{10}{*}{\rotatebox[origin=c]{90}{Citeseer}}
		& SGC & \best{$69.66 _{\pm 0.61}$} & \best{$69.56 _{\pm 1.15}$} & \secondbest{$69.06 _{\pm 1.63}$} & $67.90 _{\pm 1.37}$ \\
		& IresGCN & $65.20 _{\pm 2.21}$ & $66.30 _{\pm 1.76}$ & $66.38 _{\pm 1.46}$ & $67.26 _{\pm 2.13}$ \\
		& PSNR & $66.60 _{\pm 2.23}$ & $65.00 _{\pm 2.45}$ & $65.64 _{\pm 1.47}$ & $64.38 _{\pm 2.80}$ \\\cline{2-6}
		& GCN & $65.54 _{\pm 1.80}$ & $50.70 _{\pm 3.64}$ & $25.58 _{\pm 1.13}$ & $25.70 _{\pm 3.65}$ \\
		& GCN+LGT & \secondbest{$69.58 _{\pm 1.23}$}\improvement & \secondbest{$69.46 _{\pm 1.35}$}\improvement & \best{$69.74 _{\pm 1.07}$}\improvement & \best{$69.58 _{\pm 0.53}$}\improvement \\\cline{2-6}
		& ContraNorm & $66.52 _{\pm 1.41}$ & $66.42 _{\pm 1.07}$ & $66.34 _{\pm 1.92}$ & $66.28 _{\pm 1.57}$ \\
		& ContraNorm+LGT & $68.18 _{\pm 2.12}$\improvement & $68.48 _{\pm 1.35}$\improvement & $68.50 _{\pm 1.28}$\improvement & \secondbest{$68.52 _{\pm 1.16}$}\improvement \\\cline{2-6}
		& PairNorm & $67.66 _{\pm 2.40}$ & $67.64 _{\pm 2.12}$ & $66.66 _{\pm 2.84}$ & $65.40 _{\pm 2.00}$ \\
		& PairNorm+LGT & $69.06 _{\pm 1.90}$\improvement & $68.48 _{\pm 1.44}$\improvement & $65.12 _{\pm 2.12}$ & $65.80 _{\pm 2.50}$\improvement \\
		
		\toprule
		
		\multirow{10}{*}{\rotatebox[origin=c]{90}{Pubmed}}
		& SGC & $74.30 _{\pm 1.22}$ & $73.74 _{\pm 2.05}$ & $72.44 _{\pm 0.97}$ & $70.08 _{\pm 0.73}$ \\
		& IresGCN & $77.24 _{\pm 1.13}$ & \secondbest{$77.70 _{\pm 2.03}$} & $76.92 _{\pm 0.43}$ & \secondbest{$77.48 _{\pm 1.79}$} \\
		& PSNR & $76.94 _{\pm 1.65}$ & $77.24 _{\pm 1.33}$ & $77.24 _{\pm 0.71}$ & $77.42 _{\pm 1.07}$ \\\cline{2-6}
		& GCN & $77.04 _{\pm 2.12}$ & $70.48 _{\pm 3.43}$ & $44.24 _{\pm 3.49}$ & $47.70 _{\pm 4.11}$ \\
		& GCN+LGT & \best{$77.94 _{\pm 2.12}$}\improvement & $77.34 _{\pm 1.48}$\improvement & \secondbest{$77.82 _{\pm 1.43}$}\improvement & \best{$77.70 _{\pm 1.09}$}\improvement \\\cline{2-6}
		& ContraNorm & \secondbest{$77.94 _{\pm 2.04}$} & \multicolumn{1}{c}{OOM} & \multicolumn{1}{c}{OOM} & \multicolumn{1}{c}{OOM} \\
		& ContraNorm+LGT & $77.88 _{\pm 1.22}$ & $77.40 _{\pm 1.81}$ & \multicolumn{1}{c}{OOM} & \multicolumn{1}{c}{OOM} \\\cline{2-6}
		& PairNorm & $77.02 _{\pm 1.37}$ & \best{$77.92 _{\pm 2.33}$} & \best{$77.90 _{\pm 1.41}$} & $77.40 _{\pm 0.65}$ \\
		& PairNorm+LGT & $77.08 _{\pm 0.71}$\improvement & $76.96 _{\pm 0.94}$ & $77.14 _{\pm 0.52}$ & $77.32 _{\pm 0.98}$ \\
		
		\toprule
		
		\multirow{10}{*}{\rotatebox[origin=c]{90}{AmazonPhoto}}
		& SGC & $89.18 _{\pm 0.54}$ & $87.50 _{\pm 1.42}$ & $84.84 _{\pm 2.07}$ & $75.80 _{\pm 4.41}$ \\
		& IresGCN & $90.26 _{\pm 1.40}$ & $90.48 _{\pm 2.05}$ & $90.76 _{\pm 1.50}$ & \secondbest{$91.68 _{\pm 1.18}$} \\
		& PSNR & $90.98 _{\pm 0.61}$ & $90.72 _{\pm 1.02}$ & $91.14 _{\pm 0.31}$ & $91.26 _{\pm 1.25}$ \\\cline{2-6}
		& GCN & $90.22 _{\pm 0.61}$ & $88.40 _{\pm 0.96}$ & $64.24 _{\pm 3.18}$ & $69.94 _{\pm 5.36}$ \\
		& GCN+LGT & $91.10 _{\pm 0.76}$\improvement & $91.02 _{\pm 0.52}$\improvement & $91.26 _{\pm 0.72}$\improvement & $91.34 _{\pm 0.25}$\improvement \\\cline{2-6}
		& ContraNorm & $91.14 _{\pm 1.25}$ & \secondbest{$91.48 _{\pm 0.88}$} & \best{$91.64 _{\pm 0.45}$} & $91.56 _{\pm 0.75}$ \\
		& ContraNorm+LGT & \best{$91.74 _{\pm 0.89}$}\improvement & \best{$91.52 _{\pm 0.33}$}\improvement & \secondbest{$91.56 _{\pm 0.70}$} & \best{$91.80 _{\pm 0.91}$}\improvement \\\cline{2-6}
		& PairNorm & $91.14 _{\pm 0.59}$ & $90.54 _{\pm 0.28}$ & $90.76 _{\pm 0.53}$ & $91.00 _{\pm 1.00}$ \\
		& PairNorm+LGT & \secondbest{$91.40 _{\pm 0.71}$}\improvement & $90.94 _{\pm 0.81}$\improvement & $89.90 _{\pm 1.46}$ & $91.38 _{\pm 0.40}$\improvement \\
		\bottomrule
	\end{tabular}
	\label{master-table}
\end{table}

\textbf{Performance Gains in Deep GCNs.} Applying LGT to vanilla GCN leads to a significant accuracy boost in deep layers. On Cora, for example, GCN+LGT achieves 82.0\% accuracy at 32 layers, doubling the performance of standard GCN (41.3\%). Similar trends are observed across datasets, confirming that \textit{LGT effectively prevents over-smoothing and supports stable deep GCN training}. Additionally, GCN+LGT consistently outperforms SGC at 32 layers across all datasets, with 10–15\% gains on Cora and AmazonPhoto. Interestingly, GCN+LGT also surpasses SGC at moderate depths (4 layers) on Cora, Pubmed, and AmazonPhoto, demonstrating that \textit{proper training can balance expressiveness and smoothness}, preserving GCN’s advantages.

\textbf{Compatibility with Existing Anti-over-smoothing Methods.} When applied to ContraNorm, LGT enhances its performance on Cora, Citeseer, and AmazonPhoto. On Pubmed, where standard ContraNorm suffers from out-of-memory (OOM) issues at 8 layers, LGT improves efficiency, producing competitive results at 4 layers and valid outputs at 8 layers. For PairNorm, LGT improves performance in 10 out of 16 settings, further confirming its broad applicability to existing anti-over-smoothing techniques.

\textbf{In summary}, LGT (1) effectively mitigates over-smoothing for deep GCNs, (2) improves the deep model performance, and (3) complements existing anti-over-smoothing methods, offering a general and scalable training strategy.

\subsection{ Ablation Study}
\begin{figure}[t]
	\centering
	\includegraphics[width=0.7\textwidth]{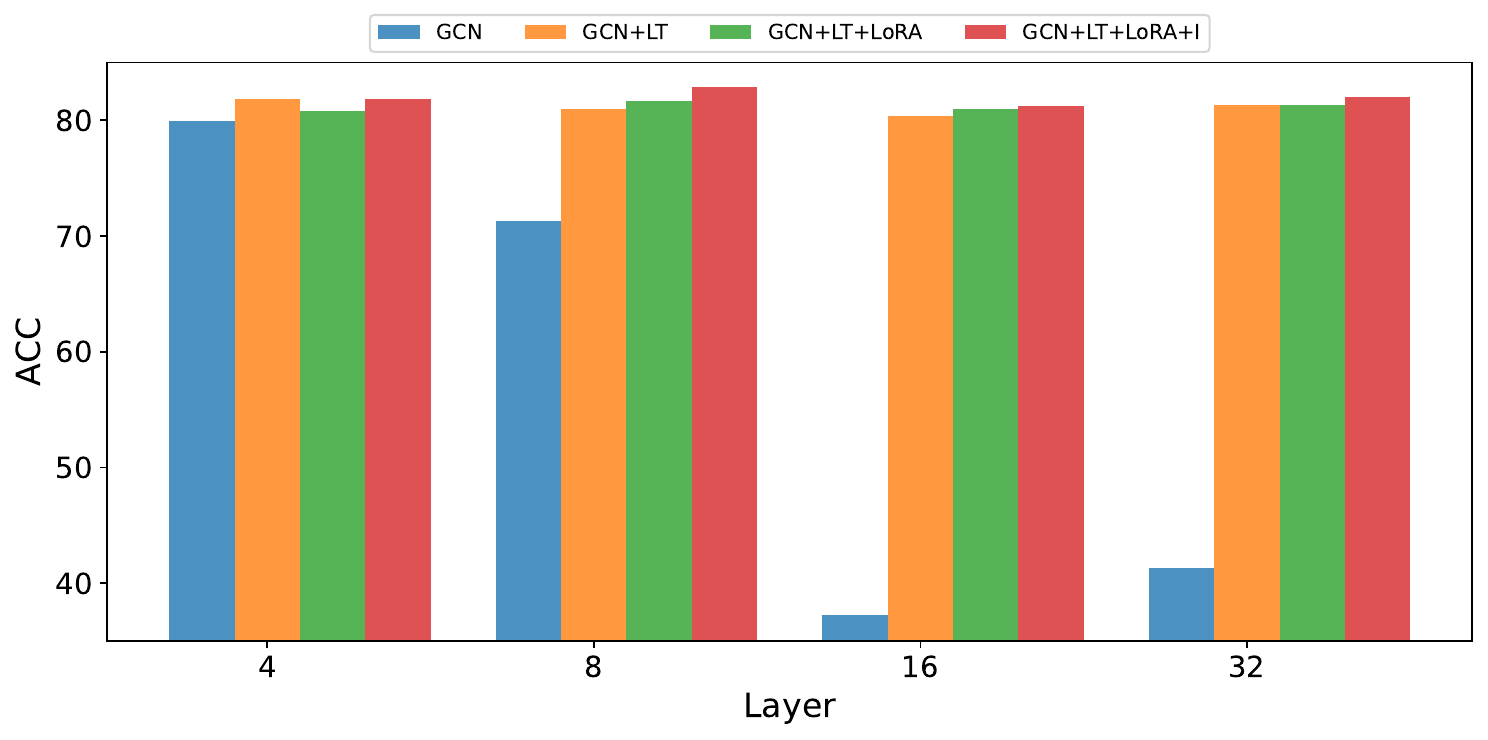}
	\caption{\textbf{Results of ablation study (Cora). }}
	\label{fig:ablation}
\end{figure}
To systematically assess the individual contributions of each component in LGT, we perform ablation studies on four GCN-based variants: (1) standard GCN as baseline, (2) GCN with Layer-wise Training (GCN+LT), (3) GCN with LT and Low-Rank Adaptation (GCN+LT+LoRA), and (4) full LGT incorporating LT, LoRA, and Identity Initialization (GCN+LT+LoRA+I). Figure~\ref{fig:ablation} reports the comparison results on the Cora dataset.

As shown in Figure~\ref{fig:ablation}, the introduction of LT significantly improves GCN performance in all depths (4 to 32 layers), with especially notable gains in 16 layers. Furthermore, adding LoRA and identity initialization brings additional improvements for deeper models (8–32 layers), demonstrating their roles in enhancing representation capacity and stabilizing optimization. These results confirm the positive and complementary contributions of all three components in alleviating over-smoothing and improving deep GCN performance.

\subsection{Training Efficiency Analysis}

\begin{figure}[t]
	\centering
	\includegraphics[width=0.9\linewidth]{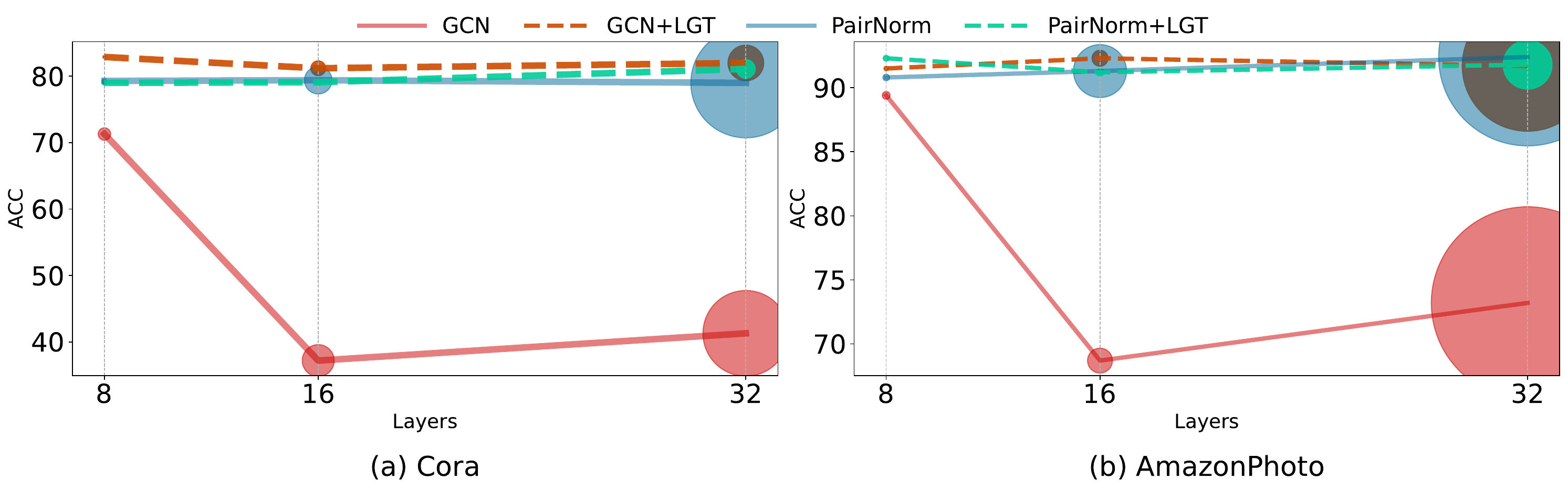}
	\caption{\textbf{Training efficiency and accuracy comparison.} Line plots show classification accuracy across different network depths, and circle sizes indicate relative training time. LGT significantly improves GCN’s accuracy while reducing training time. For PairNorm, LGT enhances training efficiency without sacrificing accuracy, highlighting its complementary effect.}
	\label{fig:efficiency}
\end{figure}

We evaluate the training efficiency of LGT on Cora and AmazonPhoto datasets, as shown in Figure~\ref{fig:efficiency}. Applying LGT to both GCN and PairNorm, we observe that LGT not only improves GCN's classification accuracy (e.g., from 70\% to over 80\% on Cora) but also significantly reduces training time by avoiding full-model optimization through progressive layer-wise updates, as shown in Table~\ref{training-time}. Although PairNorm already addresses over-smoothing, integrating LGT further reduces its training time without compromising accuracy. These results indicate that LGT complements existing normalization methods by enhancing training efficiency while maintaining or improving model performance.

\begin{table}[!t]
\centering
\caption{Training Time Comparison}
\begin{tabular}{lcccccc}
\toprule
\multirow{2}{*}{Models} & \multicolumn{3}{c}{Cora} & \multicolumn{3}{c}{AmazonPhoto} \\
\cmidrule(lr){2-4} \cmidrule(lr){5-7}
 & Layer 8 & Layer 16 & Layer 32 & Layer 8 & Layer 16 & Layer 32 \\
\midrule
GCN & 115s & 293s & 784s & 72s & 228s & 1760s \\
GCN+LGT & 28s & 135s & 326s & 41s & 1452s & 1194s \\\cline{1-7}
PairNorm & 46s & 252s & 1006s & 61s & 485s & 1627s \\
PairNorm+LGT & 42s & 62s & 170s & 55s & 69s & 447s \\
\bottomrule
\end{tabular}
\label{training-time}
\end{table}

\subsection{Effect of Rank in LoRA}
\begin{figure}[t]
	\centering
	\includegraphics[width=0.7\textwidth]{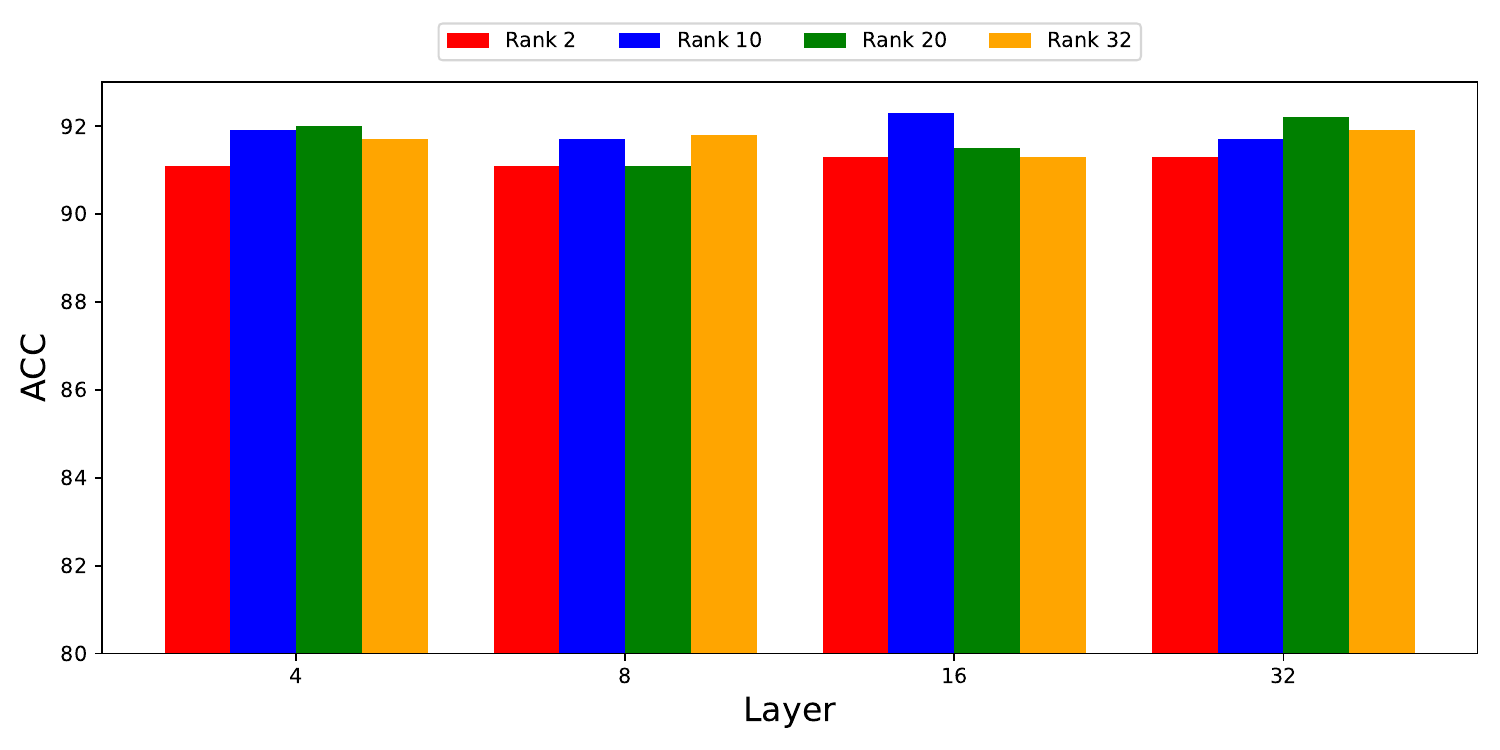}
	\caption{\textbf{Results of rank changing over layers (AmazonPhoto). }}
	\label{fig:rank}
\end{figure}
We analyze the impact of the rank parameter in LoRA on model performance under varying network depths on AmazonPhoto. As shown in Figure~\ref{fig:rank}, rank influences classification accuracy, particularly in deeper networks.

Overall, \textbf{rank=10} achieves the most stable and consistently high performance across 4, 8, and 16 layers, striking a balance between expressiveness and computational efficiency. Conversely, while \textbf{rank=32} exhibits comparatively over rank=10 in shadow networks (8 layers), it generally underperforms compared to 4 and 16 layers, likely due to over-parameterization and reduced generalization.

These results highlight the critical role of rank selection: overly small ranks limit model capacity, while excessively large ranks introduce redundancy and unnecessary computational overhead. Thus, \textbf{rank=10} is recommended for practical use, offering robust performance across depths with controlled complexity.

\subsection{Node Embedding Visualization}

\begin{figure}[h]
	\centering
	\begin{subfigure}[b]{0.40\textwidth}
		\includegraphics[width = 1\textwidth,trim=0cm 0cm 0cm 0cm, clip]{GCN_32}
		\caption{  GCN   }
	\end{subfigure}
	\begin{subfigure}[b]{0.40\textwidth}
		\includegraphics[width = 1\textwidth,trim=0cm 0cm 0cm 0cm, clip]{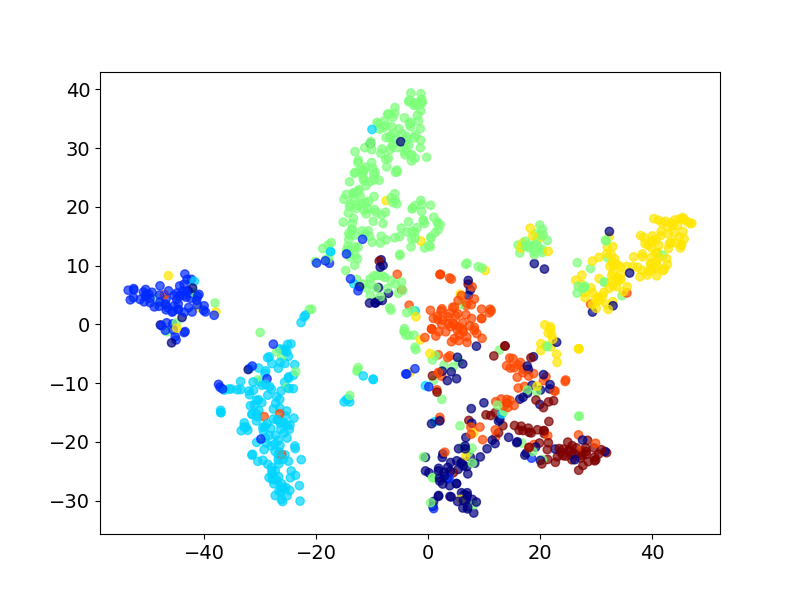}
		\caption{  GCN+LGT }
	\end{subfigure}
	\hfill
	\begin{subfigure}[b]{0.40\textwidth}
		\includegraphics[width = 1\textwidth,trim=0cm 0cm 0cm 0cm, clip]{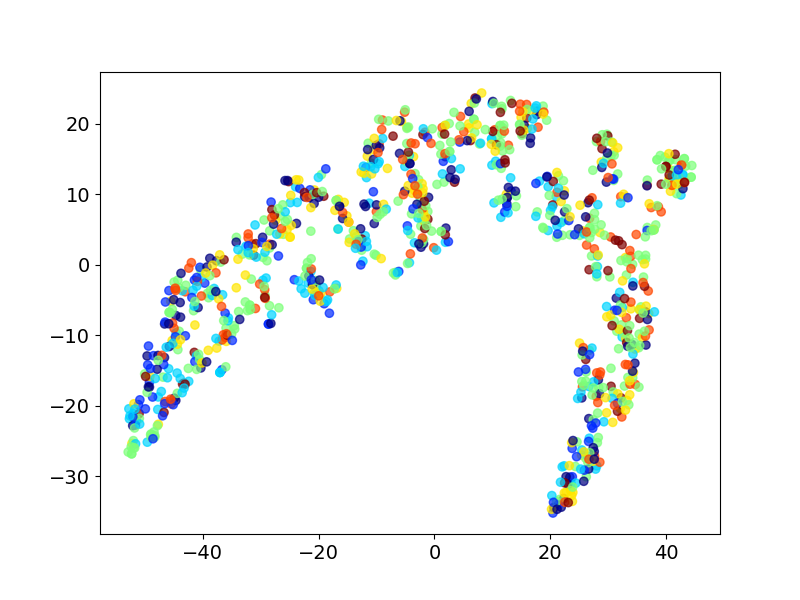}
		\caption{ContraNorm }

	\end{subfigure}
	\begin{subfigure}[b]{0.40\textwidth}
		\includegraphics[width = 1\textwidth,trim=0cm 0cm 0cm 0cm, clip]{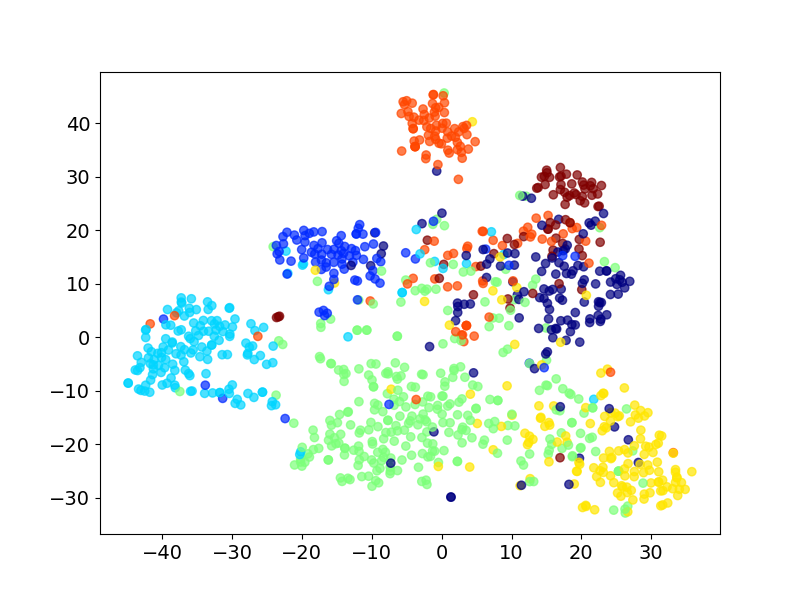}		
		\caption{ContraNorm+LGT}
	\end{subfigure}
	\hfill
	\caption{ The node embedding visualization of GCN, ContraNorm, and their respective variants GCN+LGT and ContraNorm+LGT under the 32th layer.}
	\label{fg:visualization}
\end{figure}

We visualize the 32-layer node embeddings on the Cora dataset using t-SNE to assess feature separability under deep architectures. As shown in Figure~\ref{fg:visualization}, vanilla GCN suffers from severe over-smoothing, with highly mixed node features and indistinct class boundaries. In contrast, GCN+LGT produces well-separated and compact clusters, demonstrating improved feature discrimination and effective mitigation of over-smoothing. Moreover, combining LGT with ContraNorm (ContraNorm+LGT) further enhances class separability compared to ContraNorm alone, confirming LGT's compatibility and complementary effect. These results highlight that LGT enables deep GCNs to preserve expressive and discriminative node representations.

\section{Conclusion} \label{sec:conclusion}

This paper revisits the over-smoothing problem in deep GCNs and identifies trainable linear transformations, rather than just the graph Laplacian, as a key factor exacerbating feature collapse. To address this, we propose Layer-wise Gradual Training (LGT), a novel training strategy that progressively deepens GCNs while preserving feature expressiveness. LGT integrates incremental layer-wise training for stabilized optimization, low-rank adaptation for efficient fine-tuning, and identity initialization for smooth layer expansion. Extensive experiments demonstrate that LGT significantly improves deep GCN performance, mitigating over-smoothing while enhancing training efficiency. Moreover, LGT is highly compatible with existing anti-over-smoothing techniques, such as PairNorm and ContraNorm, extending their scalability to deeper architectures. The efficiency analysis highlights faster convergence and improved stability.

Despite these advantages, two key areas warrant further exploration. (1) \textit{Theoretical insights into LGT’s optimization benefits}. The over-smoothing can be viewed as a degenerate optimization issue. LGT preserves well-initialized shallow layers and prevents unstable gradient interference from randomly initialized deep layers. A deeper theoretical analysis—possibly through saddle-point theory or local optimization landscapes—could explain how LGT helps GCNs converge toward desirable optimization regions, thereby preventing over-smoothing. (2) \textit{Interaction with existing anti-over-smoothing methods}. While LGT enhances several approaches, its impact varies across datasets (e.g., PairNorm on Pubmed). Future work will explore the relationship between training-based and structure-based regularization to refine deep GCN optimization.

Overall, LGT provides a general, training-centric solution for deep GCNs, shifting the focus from architectural design to optimization strategies and paving the way for more effective and scalable graph learning models.

\section{Acknowledgments}
This work was supported by the National Natural Science Foundation of China (Nos. 62276162, 62272286), the Fundamental Research Program of Shanxi Province (No. 202203021222016), the Central guidance for Local scientific and technological development funds (No. YDZJSX20231B001), and the Key Laboratory of Evolutionary Science Intelligence of Shanxi Province.
\bibliographystyle{}

\bibliography{}
\end{document}